\documentclass[runningheads]{llncs}
\usepackage{graphicx}

\usepackage{tikz}
\usepackage{comment}
\usepackage{amsmath,amssymb} 
\usepackage{color}
\usepackage{cite}

\usepackage{graphicx}
\usepackage{amsmath}
\usepackage{amssymb}
\usepackage{booktabs}
\usepackage{url}
\usepackage{makecell}
\usepackage{epsfig}
\usepackage{graphicx}
\usepackage[titletoc,title]{appendix}
\usepackage{comment}
\usepackage{array}
\usepackage{multirow}
\usepackage{enumitem}
\usepackage{wrapfig}
\usepackage{graphicx}
\usepackage{lipsum}
\usepackage{comment}
\usepackage{tabularx}
\usepackage{algorithm}
\usepackage{algpseudocode}
\usepackage{enumitem}
\usepackage{wrapfig}
\usepackage{hyperref}
\usepackage{authblk}

\newcommand{\topic}[1]{}
\usepackage{lipsum}

\usepackage[accsupp]{axessibility}  

\newcommand{\model}{\text{TIDEE}}
\newcommand{\Nclasses}{K}
\newcommand{\OM}{\mathcal{M}^{\textrm{O}}}

\newcommand{\maptwoD}{\mathbf{M}^{\textrm{2D}}}
\newcommand{\mapthreeD}{\mathbf{M}^{\textrm{3D}}}
\newcommand{\evisual}{\mathbf{ce}^{\textrm{vis}}}
\newcommand{\elang}{\mathbf{ce}^{\textrm{lang}}}

\newcommand{\rGCN}{\text{rGCN}}

\newcommand{\nOOP}{n_{\textrm{OOP}}}

\newcommand{\fsearch}{f_{\textrm{search}}}
\newcommand{\receptclass}{r}

\newcommand{\searchmap}{m}

\newcommand{\visualOOP}{\texttt{dDETR-OOP}}
\newcommand{\langOOP}{\texttt{BERT-OOP}}
\newcommand{\vilangOOP}{\texttt{dDETR+BERT-OOP}}
\newcommand{\oraclelangOOP}{\texttt{oracle-BERT-OOP}}

\newcommand{\oop}{\text{oop}}

\newcommand{\Memex}{\text{Memex}}
\newcommand{\threeDSemantic}{\text{\texttt{3DSmntMap2Place}}}
\newcommand{\nomemex}{\text{\texttt{WithoutMemex}}}
\newcommand{\firstRec}{\text{\texttt{RandomReceptacle}}}
\newcommand{\commonMem}{\text{\texttt{CommonMemory}}}
\newcommand{\default}{\text{\texttt{AI2THORPlacement}}}
\newcommand{\messup}{\text{\texttt{MessyPlacement}}}

\begin{document}

\pagestyle{headings}
\mainmatter
\def\ECCVSubNumber{4668}  

\title{TIDEE: Tidying Up Novel Rooms using Visuo-Semantic Commonsense Priors}

\titlerunning{TIDEE: Tidying Up with Commonsense Priors}
%
\author{Gabriel Sarch\inst{1}$^\ast$ \and
Zhaoyuan Fang\inst{1} \and
Adam W. Harley\inst{1} \and
Paul Schydlo\inst{1} \and \\
Michael J. Tarr\inst{1} \and
Saurabh Gupta \inst{2} \and
Katerina Fragkiadaki\inst{1}
}



%
\authorrunning{Sarch et al.}
%
\institute{Carnegie Mellon University \and
University of Illinois at Urbana-Champaign
}

\maketitle
{\small \centerline{$^\ast$Correspondence to \href{mailto:gsarch@andrew.cmu.edu}{gsarch@andrew.cmu.edu}}}

\begin{abstract}
We introduce $\model$, an embodied agent that tidies up a  disordered scene  based on learned commonsense  object placement  and room arrangement priors. $\model$ explores  a home environment, detects objects that are out of their natural place, infers plausible  object contexts for them, localizes such contexts in the current scene, and repositions the objects. Commonsense priors are encoded in three modules: i)  visuo-semantic  detectors that detect out-of-place objects, ii)  an associative neural graph memory of objects and spatial relations that proposes plausible semantic receptacles and surfaces for object repositions, and iii) a visual search network that guides the agent's exploration for efficiently localizing the receptacle-of-interest in the current scene to reposition the object. We test $\model$ on tidying up disorganized scenes in the AI2THOR simulation environment. $\model$  carries out the task directly from pixel and raw depth input without ever having observed the same room beforehand, relying only on priors learned from a separate set of training houses. Human evaluations on the resulting room reorganizations show $\model$ outperforms ablative versions of the model that do not use one or more of the commonsense priors. On a related room rearrangement benchmark that allows the agent to view the goal state prior to rearrangement, a simplified version of our model significantly outperforms a top-performing method by a large margin. Code and data are available at the project website: \url{https://tidee-agent.github.io/}.
 
\end{abstract}
\section{Introduction} \label{sec:intro}
\addtocounter{footnote}{-1}
\addtocounter{footnote}{-1}
\begin{figure}[t!]
    \centering
    \includegraphics[width=\textwidth]{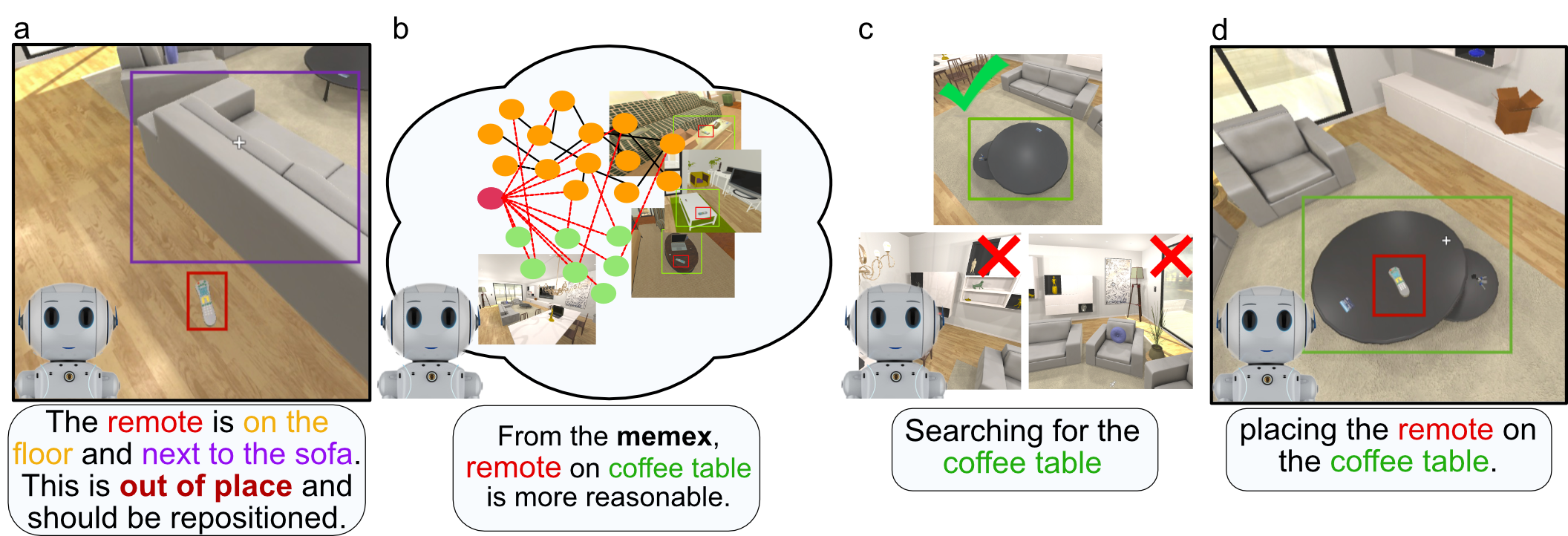}
     \caption{
    \textbf{TIDEE is an embodied agent that tidies up disorganized scenes using commonsense knowledge of object placements and room arrangements.} 
    (a) It explores the scene to detect out-of-place (OOP) objects (in this case the remote control). (b) It then infers plausible receptacles (the coffee table) through graph inference over a neural graph memory of objects and relations. (c-d) It then searches for the inferred receptacle (the coffee table) guided by a visual search network and repositions the object. 
     }
    \label{fig:teaser}
\end{figure}

\topic{we want proactive agents}
For robots to operate in home environments and assist humans in their daily lives, they need to be more than step-by-step instruction followers: they need to proactively take action in circumstances 
that violate expectations, priors, and norms, and effectively interpret incomplete or noisy instructions by human users. 
Consider Figure~\ref{fig:teaser}. A robot should  realize the remote is out-of-place,  should   be able to infer alternative plausible repositions, and tidy-up the scene by rearranging the objects to their regular locations. 
Such understanding would also permit the robot to follow incomplete instructions from human users, such as \textit{``put the remote away''}. For this, a robot needs to have commonsense knowledge regarding  contextual, object-object, and object-room spatial relations.

What is the form of this commonsense knowledge and how can it be acquired? 
There are two sources of commonsense knowledge: i) communication of such knowledge via natural language, for example, \textit{``the lamp should be placed on the bed stand"}, and ii) acquisition of such knowledge via  visually observing the world and encoding  statistical relationships between objects and places. These two sources are complementary.  Commonsense in natural language is easy to specify and modify through instruction, while commonsense through visual observation is scalable and often more expressive. Consider, for example, tall yellow IKEA lamps that are often placed on the floor, while shorter lamps are usually placed on bed stands and are appropriately centered and oriented towards the bed. In this example, object contextual relationships depend on more than the category label ``lamp''; they depend on sub-categorical information, which is easily encoded in the visual features of the objects~\cite{malisiewicz-nips09}.

We introduce Teachable Interactive Decluttering Embodied Explorer (\model), which combines  semantic and visual commonsense knowledge with embodied components  
 to tidy up  disorganized home environments it has never seen before, from raw RGB-D input. 
$\model$ explores a home environment 
to detect objects that are not in their normal locations (that therefore need to be repositioned), as shown in Figure~\ref{fig:teaser}(a). When an out-of-place (OOP) object is detected, $\model$ infers plausible  receptacles for the object to be placed onto, through graph inference over the union of a neural memory graph of objects and spatial relations and the scene graph of the room at hand (Figure~\ref{fig:teaser}(b)). It then actively explores  the scene to find instances of the predicted receptacle category guided by a visual search network, 
and repositions the detected out-of-place objects (Figure~\ref{fig:teaser}(c-d)).\footnote{We follow the terminology from AI2THOR ~\cite{ai2thor} and define a receptacle as a type of object that can contain or support other objects. Sinks, refrigerators, cabinets, and tabletops are some examples of receptacles.}  
TIDEE uses both visual features and semantic information to encode commonsense knowledge.  This knowledge is encoded in the weights of the out-of-place detectors, the neural memory graph  weights, and the visual search network weights, and is learned end-to-end to optimize objectives of the rearrangement task, such as classifying  out-of-place objects, inferring  plausible repositions, and efficiently locating an object of interest. To the best of our knowledge, this is the first work that attempts to tidy up novel 
room environments directly from pixel and depth input, without any explicit instructions 
for object placements, relying instead on learned 
prior knowledge to solve the task. 


We test $\model$ in tidying up kitchens, living rooms, bathrooms and bedrooms in the AI2THOR simulation environment~\cite{ai2thor}. We generate untidy scenes by applying random forces that push or pull objects within each room. 
We show that human evaluators prefer $\model$'s rearrangements more often than those obtained by baselines or ablative versions of our model that do not use semantics for out-of-place detection, do not use a learnable  graph memory (defaulting instead to most common placement), or do not have neural guidance during object search.  
We further show that $\model$ can be adapted to respect preferences of users by fine-tuning its out-of-place visuo-semantic object classifier based on individual instructions.   
Finally, we test a reduced version of TIDEE on the recent scene rearrangement benchmark~\cite{RoomR,Batra2020RearrangementAC}, where an AI agent is tasked to reposition the objects to bring the scene to a desirable target configuration.  TIDEE outperforms the current state of the art. 
We attribute $\model$'s excellent performance to the modular organization of its architecture and the object-centric scene representation $\model$ uses to reason about rearrangements.  


\section{Related  Work} \label{sec:related}

\subsubsection{Embodied AI.} 
The development of learning-based embodied AI agents has made significant progress across a wide variety of tasks, including: scene rearrangement~\cite{gan2021threedworld,RoomR,Batra2020RearrangementAC}, object-goal navigation~\cite{anderson2018evaluation,yang2018visual,wortsman2019learning,chaplot2020object,gupta2017cognitive,chang2020semantic}, point-goal navigation~\cite{anderson2018evaluation,savva2019habitat,wijmans2019dd,ramakrishnan2020occupancy,gupta2017cognitive}, scene exploration~\cite{chen2019learning,chaplot2020learning}, embodied question answering~\cite{gordon2018iqa,das2018embodied}, instructional navigation~\cite{anderson2018vision,shridhar2020alfred}, object manipulation~\cite{fan2018surreal,yu2020meta}, home task completion with explicit instructions \cite{shridhar2020alfred,min2021film,suglia2021embodied}, active visual learning~\cite{chaplot2020semantic,fang2020move,haber2018learning,weihs2019learning}, and collaborative task completion with agent-human conversations \cite{TEACH}. 
While these works have driven much progress in embodied AI, ours is the first agent to tackle the task of tidying up rooms, which requires commonsense reasoning about whether or not an object is out of place, and inferring where it belongs in the context of the room.
Progress in embodied AI has been accelerated tremendously through the availability of high visual fidelity simulators, such as, Habitat~\cite{savva2019habitat}, GibsonWorld~\cite{shen2020igibson},  ThreeDWorld~\cite{gan2020threedworld}, and AI2THOR~\cite{ai2thor}. Our work builds upon AI2THOR by relying on the (approximate) dynamic manipulation the simulator enables for household objects.  

\subsubsection{Representing visual commonsense.}
Visual commonsense knowledge is often represented in terms of a knowledge graph, namely, a graph of visual entity nodes (objects, parts, attributes) where edge types represent pairwise relationships between entities. 
Knowledge
graphs 
have been successfully used in visual classification and detection~\cite{marino2016more,chen2018iterative}, zero-shot classification of images~\cite{wang2018zero}, object goal navigation~\cite{yang2018visual}, and image retrieval~\cite{johnson2015image}. 

Closest to our work is the work of Yang et. al.~\cite{yang2018visual} where a  knowledge graph is used to help an agent navigate to semantic object goals. While in the knowledge graph of Yang et. al.~\cite{yang2018visual} each node stands for an object category described by its semantic embedding, in our case each node is an object instance described by both semantic and visual features, similar to the earlier work of Malisiewicz and Efros on visual $\Memex$  \cite{malisiewicz-nips09}. Moreover, we consider tidying up rooms, where navigation to semantic goals is one submodule of what the agent needs to do. Lastly, while \cite{yang2018visual} maps images to actions directly trained with reinforcement learning, and graph indexing provides simply an additional embedding to concatenate to the agent's state, our model is modular and hierarchical, using a ``theory" of out-of-place objects, inferring regular object placements, exploration to localize placements in the scene, and then taking actions to achieve the inferred object re-arrangement.   
We show that $\model$ outperforms non-modular image-to-action mapping agents in the scene re-arrangement benchmark in Section \ref{sec:scenerearrange}.

\section{Teachable Interactive Decluttering Embodied Explorer (\model)}

\begin{figure*}[t!]
    \centering
    \includegraphics[width=\textwidth]{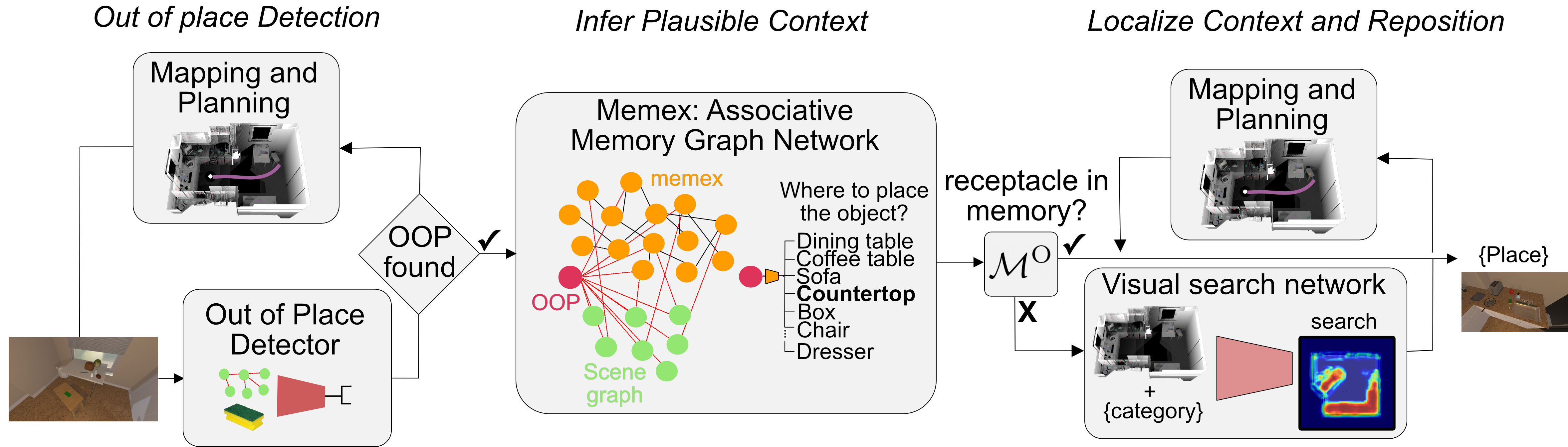}
    \caption{ \textbf{Architecture of $\model$.} 
    $\model$ explores the scene,  
    detects objects and classifies whether they are in-place or out-of-place. If an object is out-of-place, $\model$ uses graph inference in its joint external graph memory and scene graph  to infer plausible receptacle categories. It then explores the scene guided by a visual search network that suggests where instances of  a receptacle category may be found, given the scene spatial semantic map. 
    TIDEE iterates the steps above until it cannot detect any more OOP objects, in which case it concludes that the room has been tidied up.}
    \label{fig:arch}
\end{figure*}

The architecture of $\model$ is illustrated in Figure \ref{fig:arch}. The agent navigates a home environment and receives RGB-D images at each time step alongside egomotion information. We consider both groundtruth depth and egomotion, as well as noisy versions of both, and estimated depth  in our experimental section. 
The agent builds geometrically consistent spatial 2D and 3D maps of the environment by fusing RGB-D input, following prior works \cite{chaplot2020learning} (Section \ref{sec:mapping}). $\model$ detects objects and classifies them as in or out-of-place (OOP) using a combination of visual and semantic features (Section \ref{sec:OOP}). When an OOP object is detected, the agent infers plausible object context (i.e., plausible receptacle categories for the OOP object to be repositioned on) through inference over a memory graph of objects and relations ($\Memex$) and the current scene graph (Section \ref{sec:memory}). The agent then  searches  the current scene to find instances of the receptacle category and a visual search network guides its exploration by proposing locations in the scene to visit  (Section \ref{sec:activesearch}). Once the receptacle is detected, the agent places the OOP object on it. 
Navigation actions move the agent in discrete steps. For picking up and placing objects, the agent must specify an object to interact with via a relative coordinate $(x, y)$ in the (ego-centric) frame.

\subsection{Background: Semantic 3D mapping} \label{sec:mapping}

TIDEE builds 3D semantic maps of the home environment it visits augmented with 3D object detection centroids. These maps are used to infer spatial relations among objects and to guide exploration to objects-of-interest. 
Specifically, $\model$ maintains two spatial visual maps of the environment that it updates at each time step from the input  RGB-D stream, similar to previous works  \cite{chaplot2020object}: i) a 2D overhead occupancy map $\maptwoD_t \in \mathbb{R}^{H \times W}$ and, ii) a 3D occupancy and semantics map $\mapthreeD_t \in \mathbb{R}^{H \times W \times D \times \Nclasses}$, where $\Nclasses$ is the number of semantic object categories; we use $\Nclasses=116$. The $\maptwoD$ maps is used for exploration and navigation in the environment. 
More details on our exploration and planning strategy can be found in the supplementary. 

We detect objects from $\Nclasses$ semantic object categories in each input RGB image using the state-of-the-art  d-DETR detector~\cite{zhu2020deformable}, pretrained on the MS-COCO datasets~\cite{lin2014microsoft} and finetuned on images from the AI2THOR training houses. We obtain 3D object centroids by 
using the depth input to map detected 2D object bounding boxes into a 3D box centroids. 
We add these in the  3D semantic map with one channel per semantic class, similar to Chaplot et. al.~\cite{chaplot2020semantic}, but in 3D as opposed to a 2D overhead map. 
We did not use 3D object detectors directly because we found that 2D object detectors are more reliable than 3D ones likely because of the tremendous pretraining in large-scale 2D object detection datasets, such as MS-COCO \cite{lin2014microsoft}. 
Finally, to create the 3D maps $\mapthreeD$, we concatenate the 3D~occupancy maps with the 3D semantic maps .  

We further maintain an object memory $\OM$ as a list of object detection 3D centroids and their predicted semantic category labels $\OM= \{ [ (X,Y,Z)_i, \ell_i\in\{1\dots\Nclasses\} ] , i=1\dots N  \}  $, where $N$ is the number of objects detected thus far.
The object centroids are expressed with respect to the coordinate system of the agent, and, similar to the semantic maps, are updated over time using egomotion.

\subsection{Detecting out-of-place objects} \label{sec:OOP}

\begin{figure}[t!]
    \centering
    \includegraphics[width=\textwidth]{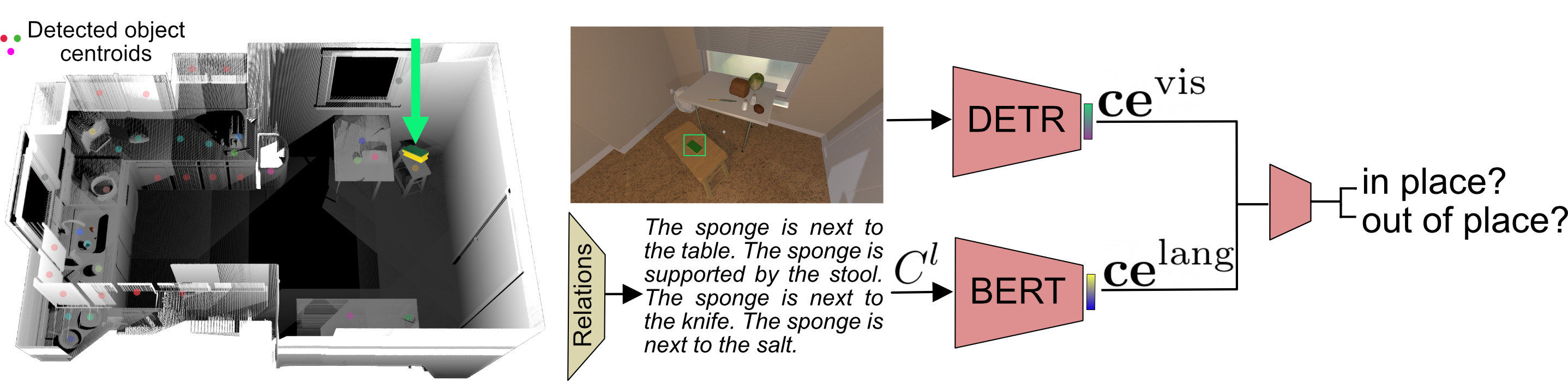}
    \caption{ \textbf{out-of-place objection classification} using  spatial language description features $\elang$ and visual features $\evisual$.}
    \label{fig:oopdet}
\end{figure}

$\model$ detects objects and classifies whether each one is in or out-of-place (OOP) 
using both visual object features 
and language descriptions  of the object's spatial relations with its surrounding objects, such as \textit{``The alarm clock is on the sofa. The alarm clock is next to the coffee table.''} We train three OOP classifiers: one that relies only on visual features, one that relies only on language  descriptions of the relations of the object with its surroundings that can more easily adapt to user preferences, and one that fuses both visual and language features, as shown in Figure \ref{fig:oopdet}. 

The visual OOP classifier ($\visualOOP$) builds upon our d-DETR detector. Specifically, we augment our d-DETR detector with a second decoding head and  jointly train it under the tasks of localizing objects and predicting their semantic categories, as well as their in or out-of-place status. We  consider the  query embedding of the d-DETR decoder as relevant visual features $\evisual$ for OOP classification. 

The language OOP classifier ($\langOOP$) infers the  relations of the detected object to surrounding objects and describes them in language form. We consider the following spatial  relations: (i) \emph{A supported-by B}, where B is a receptacle class, (ii) \emph{A next-to B}, \emph{A closest-to B}. We detect these pairwise relations using Euclidean distances on detected 3D object centroids in the object memory $\OM$. For more details on our object spatial  relation detection, please see the supplementary.  
We represent all detected pairwise relations as sentences of the form ``The \{detection class\} is \{relation\} the \{related class\}'', and concatenate the sentences to form a paragraph, as shown in Figure~\ref{fig:oopdet}. 
We map this object spatial context description paragraph into a neural vector $\elang$ for the relation set given by the [CLS] token from the BERT model~\cite{devlin2018bert} pretrained on a language masking task and then trained for plausible/non-plausible classification in our training set. 
A benefit of the language OOP classifier is that it can adapt to user's specifications  without any visual exemplars  of plausible/implausible object  arrangements.  Consider, for example, the instruction  \textit{``I want my alarm clock on the bed stand''}. Using such instruction, we generate positive and negative descriptions of in and out-of-place alarm clocks by adapting the preference into a positive sample (e.g. ``Alarm clock supported-by the bed stand''), and taking relations in the training set that include the alarm clock and a different receptacle class as negative samples (``Alarm clock supported-by the desk''). 


The multimodal classifier ($\vilangOOP$) concatenates $\evisual$ and $\elang$ as input to predict OOP classification labels for the detected object.

\begin{figure}[t!]
    \centering
    \includegraphics[width=\textwidth]{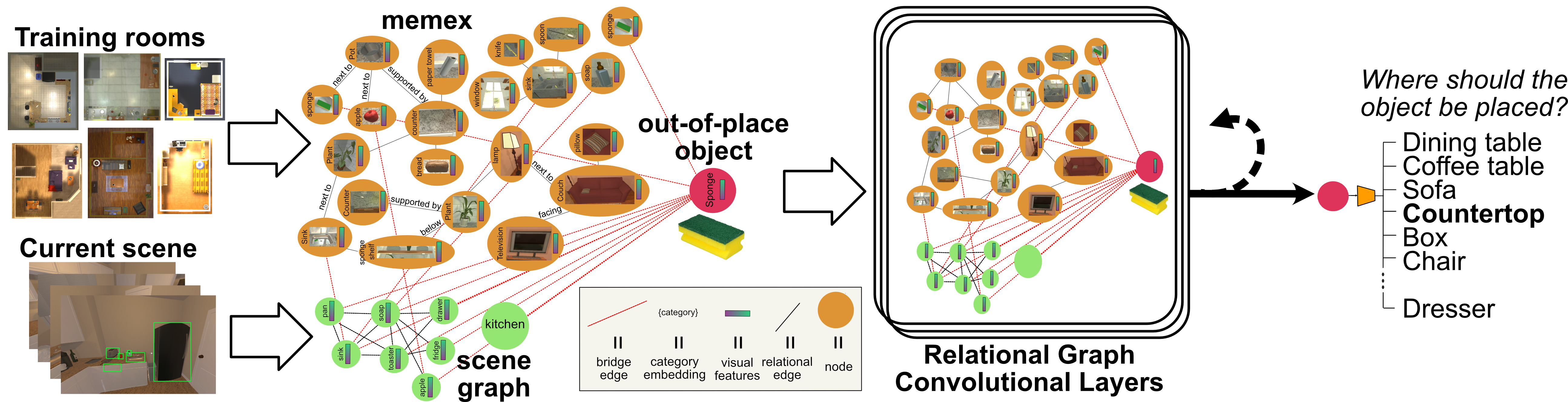}
    \caption{ \textbf{Graph inference over the union of the Memex graph and the current scene graph} infers plausible receptacle categories for an out-of-place object. }
    \label{fig:graph}
\end{figure}

\subsection{Inferring plausible object contexts with a neural associative graph memory} \label{sec:memory}

Once an OOP object is detected and picked up, $\model$ infers a plausible placement location for the object in the current scene. As shown in Figure~\ref{fig:graph}, $\model$ includes a neural graph module  which is trained to predict plausible object placement proposals of OOP objects by passing information between the OOP object to be placed, a memory graph encoding plausible contextual relations from training scenes, and a scene graph encoding the object-relation configuration in the current scene. Message passing is trained end-to-end to predict one of the possible receptacle classes in AI2THOR to place the OOP object on. 

We instantiate an OOP node, denoted $\nOOP$, consisting of the detected OOP object for which we want to infer a plausible receptacle category by concatenating the ROI-pooled detector backbone features and a category embedding of the predicted object category.

The structure of the memory graph (nodes and edges) is instantiated from 5 out of 20 training houses. Each object in the scene is given a node in the graph that consists of a category embedding and ROI-pooled detector backbone features using the bounding box of the object at a nearby egocentric viewpoint. Edge weights in the memory graph correspond to spatial relations detected between pairs of object instances that are within a distance threshold. We consider six spatial relations and corresponding edge types: \textit{above}, \textit{below}, \textit{next to}, \textit{supported by}, \textit{aligned with}, and \textit{facing}~\cite{haywardtarr95}. We infer these  using spatial relation classifiers that operate on ground-truth 3D oriented bounding boxes. 
Though the graph may contain noisy, non-important edges between object instances, for example, \textit{``the coffee table is next to the bed''} which may introduce a spurious dependence between a bed and a coffee table instance,  the  edge kernel weights are trained end-to-end  to infer plausible receptacles for  OOP objects, and thus graph inference  can learn to ignore such spurious edges.  
We call our memory graph ``$\Memex$'' to highlight that nodes represent object instances, similar to \cite{malisiewicz-nips09}, and not object categories as in previous works~\cite{yang2018visual}. 

The structure of the scene graph \cite{johnson2015image} is instantiated from observations obtained while mapping the current scene, as in Section~\ref{sec:mapping}. Nodes in the scene graph represent ROI-pooled features and category embeddings of objects detected by the agent in $\OM$. We include an additional node for the room type. We fully-connect all nodes within the scene graph. Compared to the $\Memex$ graph, we do not include separate edge weights for relations as most of the $\Memex$ relations require accurate 3D bounding boxes that we do not have access to at inference time.


We  add ``bridge edges'', as additional learnable edge weights, between nodes in the scene graph and $\Memex$ nodes with the same category, following \cite{zareian2020bridging}, to allow information to flow between the current scene and the memory graph. We further connect $\nOOP$ to all current scene nodes and to the room type node. After message passing, we pass the updated $\nOOP$ through an MLP to get logits for each possible receptacle class in AI2THOR. 

The network is trained for predicting plausible receptacles for OOP objects in 15 training houses. We use 15/20 houses to train the weights so as to not overlap with the houses used for the memory graph. Relation-specific edge weights are learned end-to-end by Relational Graph Convolutions ($\rGCN$) \cite{schlichtkrull2018modeling}. 
We supervise the network via a cross-entropy loss using ground-truth receptacle categories for each ``pickupable'' object from the AI2THOR original scene configurations. More details of our graph inference can be found in the supplementary.

\begin{figure}[t!]
    \centering
    \includegraphics[width=\textwidth]{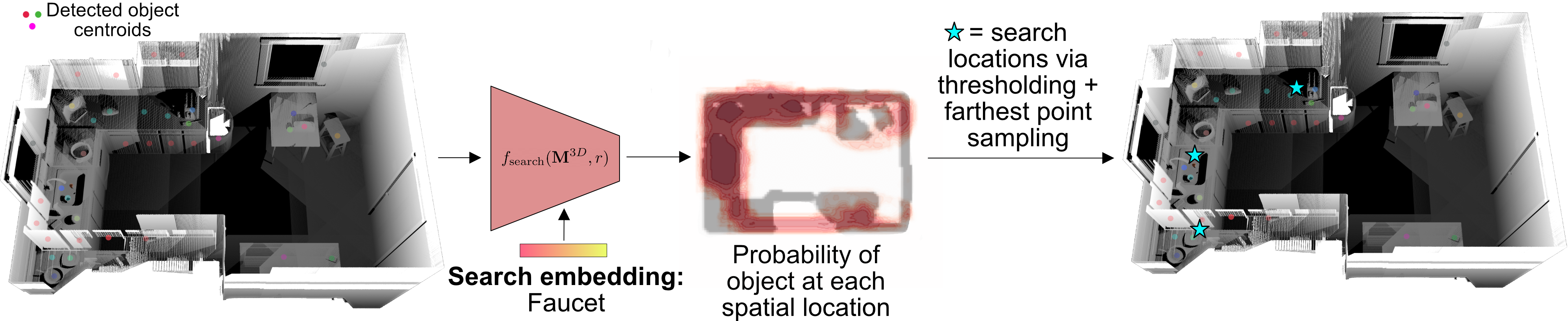}
    \caption{ \textbf{The visual search network} conditions on an object category of interest, and proposes  locations for the agent to visit  in the scene to find instances of that category.} 
    \label{fig:vsn}
\end{figure}

\subsection{Intelligent exploration using a visual  search network} \label{sec:activesearch}

After inferring a target receptacle category, $\model$ localizes it in the scene and 
 places the OOP object on top of it.  
In the  case that instances of the target receptacle category have already been detected in the scene, our agent navigates to the corresponding instance   using its navigation path planning controllers from Section \ref{sec:mapping}. In the case that the target receptacle category has not yet been detected, our model predicts plausible locations to search for the receptacle using a category-conditioned visual search network $\fsearch(\mapthreeD, r)$. 

The visual search network $\fsearch(\mapthreeD, r)$ takes as input a 3D spatial semantic map $\mapthreeD$ and a receptacle category label $\receptclass$ represented by a learned category embedding and outputs a distribution over 2D overhead locations in the current environment for $\model$ to navigate towards and find the receptacle, as shown in Figure \ref{fig:vsn}. $\fsearch$ convolves the features of the 3D semantic map with the category category features of $\receptclass$ and predicts an overhead heatmap, trained with a standard   binary cross entropy loss. 
We threshold the predicted heatmap $\searchmap$ and use non-maximum suppression via farthest point sampling to obtain a set of search locations. We rank the search locations based on their score and  visit them sequentially until the target receptacle category is detected with high probability. Further architectural details for $\fsearch$ can be found in the supplementary. 



\section{Experiments}\label{sec:experiments}
We test $\model$  on reorganizing  untidy rooms in the test  houses of the AI2THOR simulation environment.  
Our experiments aim to answer the following questions: 

(i) How well does $\model$ perform in tidying up scenes? Section~\ref{sec:replaceeval}

(ii) How much does the combination of visual and semantic features help in detecting out-of-place objects over visual features alone? Section~\ref{sec:OOPeval} 


(iii) How much does exploration guided by the proposed visual search network improve upon random exploration for detecting objects of interest? Section~\ref{sec:activesearcheval}

(iv) How well does $\model$ perform in the task of scene rearrangement~\cite{Batra2020RearrangementAC}---which requires 
memorization of a specific prior scene configuration? Section~\ref{sec:scenerearrange}

(v) How well can $\model$ adapt zero-shot to human instructions and alter placement priors accordingly? Section~\ref{sec:altpriors}

\subsection{Tidying-up task definition}

\paragraph{Dataset}
We create untidy scenes by selecting a subset of ``pickupable'' objects\footnote{Pickupable objects are a predefined set of 62 object classes in AI2THOR ~\cite{ai2thor} that are able to be picked up and repositioned by the agent, such apple, book, and laptop.}. 
We displace each object from its default location by moving the object to a random location in the scene and either dropping the object or applying a force in a random direction and allowing the AI2THOR physics engine to resolve the object's end location. We consider all available room types, namely bedrooms, living rooms, kitchens and bathrooms. We generate 8000 training, 200 validation, and 100 testing messy configurations. The goal of the agent is to manipulate the messy objects back to plausible locations within the room. An episode ends once the agent executes the ``done'' action or a maximum of 1000 steps have been taken. For more details on the task and dataset, please see the supplementary. 

 
\subsection{Object repositioning evaluation} \label{sec:replaceeval}

We have $\model$ and all baselines perform the tidy task to detect out-of-place objects and reposition them within the scene.

\paragraph{Evaluation metrics} 
 Quantitative evaluation of object repositioning is difficult: an object may have multiple plausible locations in a scene, and therefore measuring the distance from a single initial ground-truth 3D location is usually not reflective of performance. We thus evaluate the plausibility of object repositions of our model from those of baseline models by querying human evaluators in Amazon Mechanical Turk (AMT). Given two candidate repositions by for the same object $\model$ and a baseline, we ask human evaluators to select the one they find most plausible. 
We include the AMT interface we used in the supplementary.

\begin{table}[h]
\parbox{0.55\linewidth}{
\centering
\caption{\textbf{Percent of human evaluators that prefer $\model$ object repositions versus baselines.} Reported is mean and standard error across subjects (n=5). All preferences are significantly above chance (*p$<$0.05, **p$<$0.01, Binomial test). Bold indicates higher preference for $\model$.}\label{tab:preference}
\begin{tabular}{lccc}
\toprule
$\model$ vs $\commonMem$ & \textbf{54.30$\pm$3.32}*\\
$\model$ vs $\nomemex$ & \textbf{54.32$\pm$4.67}*\\
$\model$ vs $\threeDSemantic$ & \textbf{ 57.69$\pm$1.29}**\\
$\model$ vs $\firstRec$ & \textbf{ 64.59$\pm$2.94}**\\
$\model$ vs $\messup$ & \textbf{ 92.06$\pm$1.57}**\\
$\model$ vs $\default$\text{ } & 34.00$\pm$3.13**\\
\bottomrule
\end{tabular}
}
\hfill
\parbox{.4\linewidth}{
\centering
\caption{\textbf{Evaluating visual search performance for finding objects of interest in test scenes} for $\model$ and an exploration baseline that uses our 2D overhead occupancy maps to propose random search locations~\cite{yamauchi1997frontier}.}
\label{tab:vsnevaluation}
\begin{tabular}{@{}ccc@{}}
\toprule
 &  \% Success $\uparrow$ & Time Steps $\downarrow$  \\ 
 \midrule
 $\model$   & \textbf{72.4} &  \textbf{88.8} \\
 w/o VSN   & 64.8 &  100.9\\
\bottomrule
\end{tabular}
}
\end{table}

\paragraph{Baselines} 
We compare $\model$ against baselines that vary in their way of inferring plausible receptacle categories for repositioning of out-of-place  objects. 
All baselines use the same mapping and planning for navigation, the same multimodal classifier  for detecting out-of-place objects ($\vilangOOP$), and the visual search network for localizing receptacle instances of a category. 
We compare placements from $\model$ against the following baselines:  
(i) $\commonMem$: A model that  considers the most common receptacle in the training set for the out-of-place object category. 
(ii) $\nomemex$: A model that uses the scene graph but not the $\Memex$ for graph inference. 
(iii) $\threeDSemantic$: A model that  proposes  
 repositioning locations within the current scene by conditioning the visual search network on the category label of the out-of-place object. We threshold all predicted map locations and do farthest point sampling to obtain a set of diverse object placement proposals. The proposals are sorted by confidence value and visited sequentially until any receptacle is found within the local region of the proposed location. 
(iv) $\firstRec$: A model that selects as the target receptacle the first receptacle detected by a random exploration agent. 
(v) $\default$: The  location of the OOP object in the original (tidy) AITHOR scene. The default object positions  usually follow commonsense priors of scene arrangements. 
(vi) $\messup$:  The location of the OOP object in the messy scene. 

We report human preferences for OOP object repositions for our model versus each of the baselines in Table~\ref{tab:preference}. $\model$ is preferred 54.3\% of the time over $\commonMem$, the most competitive of the baselines. $\commonMem$ does not consider the visual features of the out-of-place object, rather, only its semantic category, and thus cannot reason using sub-categorical information regarding object placements. $\model$ is still preferred 34\% of the time over the $\default$ placements indicating that its re-placements are plausible and competitive with an oracle. We note that 
a perfect model would at best obtain a (50-50) preference compared to these placements provided by the AITHOR environment designers. 






\subsection{Out-of-place detector evaluation} \label{sec:OOPeval} 
In this section, we evaluate TIDEE's accuracy for detecting objects in and out-of-place from images collected from the test home environments. An in-place object is one in its default location in the AITHOR scene, while an out-of-place object is one moved out-of-place as defined at the beginning of Section \ref{sec:experiments}. We compute  average precision (AP) at IOU thresholds of 0.25 and 0.5 for in-place (\textit{IP}) and out-of-place (\textit{OOP}) objects, as well as the meanAP (mAP) for 
visual only ($\visualOOP$), language only ($\langOOP$) and multimodal ($\vilangOOP$) classifiers described in Section \ref{sec:OOP}.  
We also compare against an oracle BERT classifier that assumes access to ground-truth 3D object centroids, bounding boxes, and category labels to detect relations and form descriptive utterances of in and out-of-place objects, which we call $\oraclelangOOP$.

We show quantitative comparisons in Table~\ref{tab:oop_supp}. Combining language and visual features performs slightly better than using language or visual features alone for out-of-place object detection. The benefit of the language classifier is that it can be re-trained on-demand to adjust to human instructions without any visual training data, as we explain in Section \ref{sec:altpriors}. 
The good performance of the oracle BERT classifier suggests that simple relations inferred from accurate 3D centroids likely suffice to classify in- and out-of-place objects in AI2THOR scenes if perception is perfect. 

\begin{table} 
\begin{center}
\caption{\textbf{Average precision (AP) for in and out-of-place object detection. }
Combining vision and language features helps detection performance. IP = in place; OOP = out of place. 
}\label{tab:oop_supp}
\begin{tabular}{@{}ccccccc@{}}
\toprule
 & mAP$_{0.25}$ & AP$_{0.25}^{IP}$ & AP$_{0.25}^{OOP}$ & mAP$_{0.5}$ & AP$_{0.5}^{IP}$ & AP$_{0.5}^{OOP}$\\ 
 \midrule
$\vilangOOP$ & \textbf{51.09} & \textbf{58.41} & \textbf{43.78} & \textbf{46.26} & \textbf{53.64} & \textbf{38.88} \\
$\visualOOP$ & 49.98 & 57.60 & 42.37 & 44.98 & 52.79 & 37.17\\
$\langOOP$ & 31.71 & 41.13 & 22.30 & 25.25 & 33.79 & 16.71\\
\hline
$\oraclelangOOP$ & -- & -- & -- & 90.70 & 96.24 & 85.16 \\
\bottomrule
\end{tabular}
\end{center}
\end{table}

\subsection{Visual search network evaluation} \label{sec:activesearcheval}
In this section, we compare exploration for finding objects of interest in test scenes (one category of the possible 116 per episode) guided by \model{}'s visual search network against an exploration agent that uses the 2D overhead occupancy map and samples unvisited locations to visit, similar to Yamauchi~\cite{yamauchi1997frontier}. We adopt the success criteria similar to the object goal navigation~\cite{batra2020objectnav} and define a successful trial as one where the agent is within a radius of \textit{any} target object category instance and the object is visible within view. We report the percentage of successful episodes performed by the agent and average number of time steps across all episodes in Table \ref{tab:vsnevaluation}. If an agent fails an episode, the number of time steps defaults to the maximum allowable steps for each episode (200). 
\model{} outperforms the exploration baseline. We show visualizations of the network predictions in  Figure~\ref{fig:vissearch}, and also in the supplementary.

\begin{figure*}[t!]
    \centering
    \includegraphics[width=\textwidth]{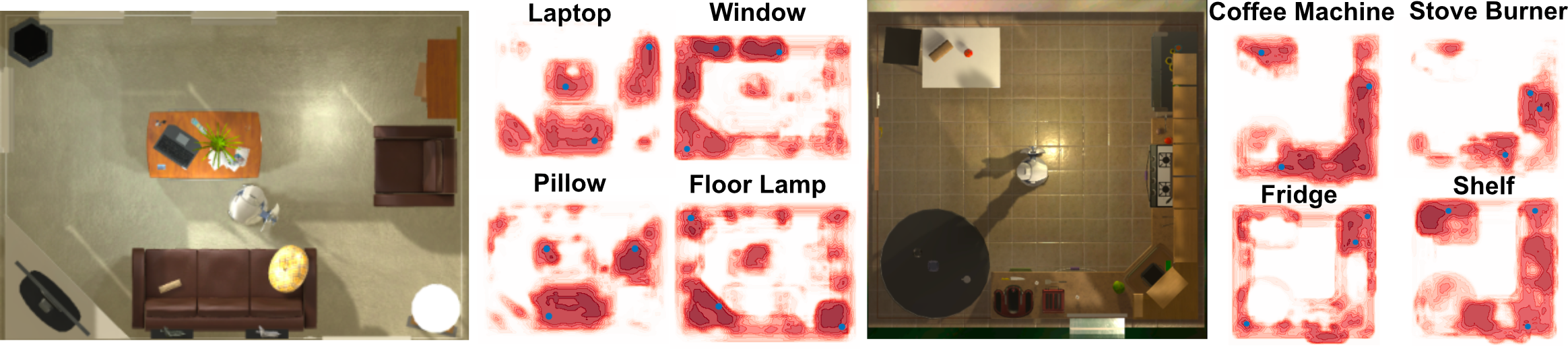}
    \caption{ \textbf{Visual Search Network} predictions encode  object location priors for different object categories.} 
    \label{fig:vissearch}
\end{figure*}


\subsection{Scene Rearrangement Challenge} \label{sec:scenerearrange}
We test $\model$ to generalize to the recent scene rearrangement benchmark of \cite{RoomR}, 
which considers an AI agent tasked with repositioning objects in a scene in order to  match the prior configuration of an identical scene.  
We consider the two-phase rearrangement setup where in the first ``walkthrough" phase, the agent observes a room in its initial configuration, and in the second so called ``unshuffle" phase, observes the same room with some objects in new configurations and is tasked to rearrange the room back to its initial configuration. 
While the challenge considers both rearranging objects to different locations  within a room and changing their open/close states, we only consider repositioning of objects 
because our current model does not handle opening and closing receptacles. 

We simplify $\model$'s architecture and only maintain the 2D \& 3D occupancy map for navigation and the object memory $\OM$ for keeping track of objects and their labels over time. We start each phase by exploring the scene and detecting objects. As in Section \ref{sec:OOP}, we infer the relations for all pickupable objects in the object memory $\OM$ in the initial and shuffled scenes. We consider an object of the initial scene displaced if its category label has been detected in the shuffled scene and the proportion of inferred relations that are different across the two scenes (\{\# same relations\}/\{\# different relations\}), initial and shuffled, is less than a threshold (we use 0.35). 
For example, a bowl with relations \textit{bowl next to sink, bowl supported by countertop, bowl next to cabinet} in the initial scene, and relations \textit{bowl next to chair, bowl supported by dining table, bowl next to lamp} in the shuffled scene is considered misplaced by $\model$. 
Then, our agent  navigates  to the object's 3D location detected in the initial scene and places it there. Our agent uses the navigation controllers from Section~\ref{sec:mapping}. 

We use the evaluations metrics described in Weihs et. al.~\cite{RoomR} : (1) Success ($\uparrow$): the trial is a success if the initial configuration is fully recovered in the unshuffle phase; (2) \% FixedStrict ($\uparrow$): the proportion of objects that were misplaced initially but ended in the correct configuration (if a single in-place object is moved out-of-place, this metric is set to 0); (3) \% Energy ($\downarrow$): the energy is a measure for the similarity of the rearranged scene and the original scene, the lower the more similar (for more details, refer to Weihs et. al.~\cite{RoomR}); (4) \% Misplaced ($\downarrow$): this metric equals the number of misplaced objects at the end of the episode divided by the number of misplaced objects at the start.

We report $\model$'s performance compared to the top performing methods for the two-phase re-arrangement in Table~\ref{tab:rear}. The model from Weihs et. al.~\cite{RoomR} trains a reinforcement learning (RL) agent with proximal policy optimization (PPO) and imitation learning (IL) given RGB images as input and includes a semantic mapping component adapted from the Active Neural SLAM model~\cite{chaplot2020learning}. 
We additionally show the robustness of $\model$ to realistic sensor measurements. We consider three different versions of \model{} depending on the source of egomotion and depth information: \textbf{(i)} TIDEE uses ground-truth egomotion and depth. \textbf{(ii)}  TIDEE\textit{+noisy pose} uses ground-truth depth and egomotion from the LocoBot agent in AI2THOR with Gaussian movement noise added to each movement based on measurements of the real LocoBot robot~\cite{murali2019pyrobot} (forward movement $\sigma$ = 0.005 meters; rotation $\sigma$ = 0.5 degrees). \textbf{(iii)} TIDEE\textit{+est. depth} uses ground-truth egomotion and depth obtained from  the depth prediction model of Blukis et. al.~\cite{blukis2022persistent}, which takes in egocentric RGB images. The model is pre-trained and then finetuned on the training scenes of ALFRED~\cite{shridhar2020alfred}.

\subsection{Updating placement priors by instruction} \label{sec:altpriors}
In this section, we test whether we can alter the OOP classifier on-demand using language specifications  for in and out-of-place. 
Since alarm clocks are often found on desks in AI2THOR, 
we tested whether augmenting training by pairing the sentence \textit{``alarm clock is supported by desk"} with the out-of-place label would allow us to alter the OOP classifier's output. 
As shown in Table~\ref{tab:oop_prior}, across three test scenes where alarm clocks are found on desks, the initial OOP object classifier gives us low probability that the alarm clock on the desk is out-of-place. We then add in the language description  \textit{``alarm clock is supported by desk"} for a small amount of additional iterations. As shown in Table~\ref{tab:oop_prior}, we find that our procedure suffices to alter the priors of the classifier. 
We provide additional examples using various object-relation pairings in the supplementary.

\begin{table}[t] 
\begin{center}
\caption{\textbf{Test set performance on 2-Phase Rearrangement Challenge (2022).} $\model$ outperforms the baseline of \cite{RoomR} even with realistic noise.}\label{tab:rear}
\begin{tabular}{@{}lcccc@{}}
\toprule
& \%  FixedStrict $\uparrow$ & \% Success $\uparrow$ & \% Energy $ \downarrow$ & \% Misplaced $\downarrow$\\
 \midrule
 $\model$ & \textbf{11.6} & \textbf{2.4} & \textbf{93} & \textbf{94} \\
 $\model$ \scriptsize\textit{+noisy pose} & 7.7 & 1.2 & 101 & 101 \\
 $\model$ \scriptsize\textit{+est. depth} & 5.9 & 0.6 & 97 & 97 \\
 $\model$ \scriptsize\textit{+noisy depth} & 11.4 & 2.0 & 94 & 95 \\
 Weihs \textit{et al.}~\cite{RoomR} & 0.5 & 0.0 & 110 & 110 \\
\bottomrule
\end{tabular}
\end{center}
\end{table}


\begin{wraptable}[14]{r}{4.0cm}
\centering
\caption{\textbf{Altering priors with instructions.} 
The confidence of the out-of-place classifier for clocks found on desks in three test scenes increases when the additional spatial description for indicating out-of-place clocks.\\}\label{tab:oop_prior}
\begin{tabular}{@{}lccc@{}}
\toprule
& Before & After \\ \midrule
Clock \#1  & .08  & .73     \\
Clock \#2 & .10 & .62    \\
Clock \#3 & .12 & .76 \\
\bottomrule
\end{tabular}
\end{wraptable}\paragraph{Limitations.} \label{sec:limitations}
$\model$ has the following two limitations: 
i) It does not consider open and closed states of objects, or their 3D pose as part of the messy and reorganization process, which are direct avenues for future work.
ii) The messy rooms we create by randomly misplacing objects may not match the messiness in human environments. 


\section{Conclusion} \label{sec:lim}
We have introduced $\model$, an agent that tidies up rooms in home environments using commonsense priors encoded  in  visuo-semantic out of place detectors, visual search  networks that guide exploration to objects, and a Memex neural graph memory of objects and  relations that infers plausible object context. 
We evaluate with human evaluators, and find that $\model$ outperforms agents that lack it's modular architecture, as well as  modular agents that lack $\model$'s commonsense priors. 
$\model$ can be instructed in natural language to follow on-demand specifications for object placement. 
Finally, we establish a new state-of-the-art for the scene rearrangement challenge of Weihs et. al.~\cite{RoomR} by simplifying $\model$'s architecture to memorize a single scene as opposed to using a prior learned across multiple environments. We believe $\model$ takes an important step towards embodied visuo-motor commonsense  reasoning. 
\\\

\noindent\textbf{Acknowledgements.} \label{sec:ack}
This material is based upon work supported by National Science Foundation grants GRF DGE1745016 \& DGE2140739 (GS), a DARPA Young Investigator Award, a NSF CAREER award, an AFOSR Young Investigator Award, and DARPA Machine Common Sense. Any opinions, findings and conclusions or recommendations expressed in this material are those of the authors and do not necessarily reflect the views of the United States Army, the National Science Foundation, or the United States Air Force.

\clearpage



%
%
\bibliographystyle{splncs04}
\bibliography{egbib,7_refs_new}
\clearpage
\makeatletter
\renewcommand \thesection{S\@arabic\c@section}
\renewcommand\thetable{S\@arabic\c@table}
\renewcommand \thefigure{S\@arabic\c@figure}
\renewcommand \thealgorithm{S\@arabic\c@algorithm}
\makeatother

\setcounter{section}{0}
\setcounter{figure}{0}  
\setcounter{table}{0} 

\renewcommand{\theHsection}{Supplement.\thesection}
\renewcommand{\theHtable}{Supplement.\thetable}
\renewcommand{\theHfigure}{Supplement.\thefigure}

\section{Overview}
Section~\ref{sec:id} contains more details of the methods described in the main paper. Section~\ref{sec:exp_det} provides additional details on the experiments. Section~\ref{sec:add_exp} provides additional evaluation of the networks. 

\section{Implementation details} \label{sec:id}

\subsection{Virtual environment and action space}
We use the following actions: move forward, rotate right, rotate left, look up, look down, pick up, put down. We rotate in the yaw direction by 90 degrees, and rotate in the pitch direction by 30 degrees. We do not constrain our agent to grid locations. The RGB and depth sensors are at a resolution of 480x480, a field of view of 90 degrees, and lie at a height of 0.9015 meters. The agent's coordinates are parameterized by a single $(x,y,z)$ coordinate triplet with $x$ and $z$ corresponding to movement in the horizontal plane and $y$ reserved for the vertical direction. Picking up objects occurs by specifying an (x,y) coordinate in the agent's egocentric frame. If by ray-tracing, the point intersects an object that is pickupable and within 1.5 meters of the agent, then the pickup action succeeds. Placing objects occurs by specifying an (x,y) coordinate in the agent's egocentric frame to place the object. If by ray-tracing, the point intersects an object that is a receptacle class, has enough free space in the radius of the target location, and within 1.5 meters of the agent, then the place action succeeds if the agent is holding an object. 
Since some objects require their state to be open for placement to successfully occur (e.g. Fridge), the agent will also try to open the receptacle if placement initially fails. 

\subsection{Pseudo code for $\model$}
We present pseudo code for the $\model$ algorithm in Algorithm~\ref{alg:cap}. We denote FMM to mean Fast Marching Method~\cite{sethian1996fast}, g to denote the point goal in the 2D overhead map $\maptwoD$, $r$ to denote a receptacle, and fps to denote farthest point sampling. If $\model$ does not find one of the predicted receptacles from the rGCN network, $\model$ will attempt to retrieve a general receptacle class from its memory of detected objects, navigate there, and attempt to place it. If after $m$ placement attempts the object is still not placed successfully (for example if $\model$ gets stuck while navigating), $\model$ will drop the object at its current location and resume the out-of-place search.  

\begin{algorithm}
\caption{$\model$ algorithm}\label{alg:cap}
\begin{algorithmic}
\While{$unexplored\_area > A$}  \Comment{Mapping the scene}
    \If{g reached}
    \State Sample new g in unexplored area
    \EndIf
    \State Execute movement with FMM to g
    \State Update $\maptwoD$, $\mapthreeD$, $\OM$
\EndWhile
\State Sample new g in reachable area \Comment{out-of-place detection}
\While{not $oop$ found after sampling $k$ goals} 
\If{g reached}
\State Sample new g in reachable area
\EndIf
\State Execute movement with FMM to g
\State Update $\maptwoD$, $\mapthreeD$, $\OM$
\State Run $\vilangOOP$
\If{$\oop$ found} 
\State navigate to $\oop$, Execute PickupObject
\State $\receptclass$ $\gets$ Run $\rGCN$ \Comment{Infer plausible context}
\If{$\receptclass \in \OM$}
\State navigate to $\receptclass$ with FMM, Execute PutObject
\Else
\State $\searchmap$ $\gets$ Run $\fsearch$ \Comment{Localize context}
\For{g $\in$ fps($\searchmap$)}
\State navigate to g with FMM
\If{$\receptclass$ detected}
\State navigate to $\receptclass$ with FMM
\State Execute PutObject
\EndIf
\EndFor
\EndIf
\EndIf
\EndWhile
\end{algorithmic}
\label{alg:tidee}
\end{algorithm}

\subsection{Semantic mapping and planning}

$\model$ maintains two spatial visual maps of its environment that it updates at each time step from the input  RGB-D stream: i) a 2D overhead occupancy map $\maptwoD_t \in \mathbb{R}^{H \times W}$ and, ii) a 3D occupancy and semantics map $\mapthreeD_t \in \mathbb{R}^{H \times W \times D \times \Nclasses}$, where $\Nclasses$ is the number of semantic object categories, we use $\Nclasses=116$. The $\maptwoD$ maps are used for exploration and navigation in the environment. The $\mapthreeD$ maps are used for inferring locations of potential receptacles conditioned on their semantic categories, as described in Section 3.4 of the main paper.  

At every time step $t$, we unproject the input depth maps using intrinsic and extrinsic information of the camera to obtain a 3D occupancy map registered to the coordinate frame of the agent, similar to earlier navigation agents \cite{chaplot2020learning}. The 2D overhead maps $\maptwoD_t$ of obstacles and free space are computed by projecting the 3D occupancy along the height direction at two height levels and summing.
For each input RGB image, we run a state-of-the-art d-DETR detector \cite{zhu2020deformable} (pretrained on COCO~\cite{lin2014microsoft} then finetuned on AI2THOR) to localize each of $\Nclasses$ semantic object categories. 
Similarly, we use the depth input to map detected 2D object bounding boxes into a 3D centroids dilated with Gaussian filtering and add them into the  3D semantic map, we have one channel  per semantic class---similar to \cite{chaplot2020semantic}, but in 3D as opposed to a 2D overhead map. 
We did not use 3D object detectors directly because we found that 2D object detectors are more reliable than 3D ones simply because of the tremendous pretraining in large-scale 2D object detection datasets, such as MS-COCO \cite{lin2014microsoft}. 
Finally, 3D maps $\mapthreeD$ result from the concatenation of the 3D~occupancy maps with the 3D semantic maps.  
Alongside the 3D semantic map $\mapthreeD$, we maintain an object memory $\OM$ as a list of object detection 3D centroids and their predicted semantic labels $\OM= \{ [ (X,Y,Z)_i, \ell_i\in\{1...\Nclasses\} ] , i=1..K  \}  $, where $K$ is the number of objects detected thus far.
The object centroids are expressed with respect to the coordinate system of the agent, and, similar to the semantic maps, updated over time using egomotion.

\paragraph{Exploration and path planning} 

 $\model$ explores the scene using a classical mapping method. We take the initial position of the agent to be the center coordinate in the map. We rotate the agent in-place and use the observations to instantiate an initial map. Second, the agent incrementally completes the maps by randomly sampling an unexplored, traversible location based on the 2D occupancy map built so far, and then navigates to the sampled location, accumulating the new information into the maps at each time step. The number of observations collected at each point in the 2D occupancy map is thresholded to determine whether a given map location is explored or not. 
Unexplored positions are sampled until the environment has been fully explored, meaning that the number of unexplored points is fewer than a predefined threshold.

To navigate to a goal location, we compute the geodesic distance to the goal from all map locations using a fast-marching method ~\cite{sethian1996fast} given the top-down occupancy map $\maptwoD$ and the goal location in the map. We then simulate action sequences and greedily take the action sequence which results in the largest reduction in geodesic distance.

\subsection{2D-to-3D unprojection}\quad For the $i$-th view, a 2D pixel coordinate $(u,v)$ with depth $z$ is unprojected and transformed to its coordinate $(X,Y,Z)^T$ in the reference frame:
\begin{equation}
    (X,Y,Z,1) = \mathbf{G}_{i}^{-1} \left(z \frac{u-c_{x}}{f_{x}}, z \frac{v-c_{y}}{f_{y}}, z, 1\right)^{T}
\end{equation}
where $(f_x, f_y)$ and $(c_x, c_y)$ are the focal lengths and center of the pinhole camera model and $\mathbf{G}_i \in SE(3)$ is the camera pose for view $i$ relative to the reference view. This module unprojects each depth image $I_i \in \mathbb{R}^{H\times W \times3}$ into a pointcloud in the reference frame $P_i \in \mathbb{R}^{M_i \times 3}$ with $M_i$ being the number of pixels with an associated depth value. 

We voxelize the point cloud into a 128x64x128 occupancy $\in \{0,1\}$ centered at the initial position of the agent, and aggregate (take max) the occupancies across views to obtain $M_t^o \in \{0,1\}$.

\subsection{Object tracking and semantic aggregation.} \label{sec:obj_track}
As described in Section 3.2, we track previously detected objects by their 3D centroid $C \in \mathbb{R}^{3}$. We estimate the centroid by taking the 3D point corresponding to the median depth within the bounding box detection and bring it to a common coordinate frame. We extend previous work \cite{chaplot2020semantic} to 3D and add a channel to the 3D occupancy map for each object category. For each detected centroid $C^j$ of class index $j$, we accumulate it into a 3D occupancy map. We then apply a Guassian filter $g$ to dilate the centroids in the map and add this to to the $jth$ channel of the 3D semantic occupancy map $M_t$. Thus, the $jth$ channel of the 3D semantic map at time step $t$ can be written as: 
\begin{equation}
M_t^j = M_t^o + g(f(C^j))
\end{equation}
where $M_t^o \in \mathbb{R}^{H  \times W  \times D}$ is the accumulated 3D occupancy, $g$ is a guassian filter operation, and $f$ accumulates each centroid $i$ in class index $j$ into an occupancy map $M \in \mathbb{R}^{H  \times W  \times D}$. Centroids are more robust to noisy depth and detection estimates, and often provide enough information for active search and object spatial tracking.

\subsection{Out-of-place detector} \label{sec:oop_details}
As described in Section 3.2 of the main paper, our OOP detector makes use of visual and relational language as input to our OOP network. We generate training scenes with some objects out-of-place using the same algorithm described in Section~\ref{sec:messup}. We first finetune deformable-DETR~\cite{zhu2020deformable} (pretrained on COCO~\cite{lin2014microsoft}) on the training houses (object seed randomized) to predict the bounding boxes, semantic segmentation masks, and semantic labels by generating random trajectories through the scene. We then train on the messup configurations and add an additional classification loss on the output decoder queries to predict whether the object is in- or out-of-place. We use the output decoder queries for the $\visualOOP$ classifier. 

For the language detector, we freeze the detector described above, and use it to update our object tracker $\OM$ while the agent explores the scene. Then, the agent visits a location to search for an out-of-place object and for each object detected in view above a confidence threshold, we infer its relations described in Section~\ref{sec:rel_cent} with all objects in memory, and systematically combine them into a paragraph of text. An example paragraph is shown below. 
\small
\emph{The pillow is next to the key chain. The pillow is next to the laptop. The pillow is next to the side table. The pillow is next to the mug. The pillow is next to the teddy bear. The pillow is supported by the side table. The pillow is closest to the mug.}
We make use of the extensive pretraining of the BERT language model~\cite{devlin2018bert} as a starting point for our language classifier. We tokenize the paragraph text and give it as input to the BERT model. For the language-only detector ($\langOOP$), we give the pooled output \{cls\} token from BERT to a three-layer fully-connected classifier to predict in or out-of-place. 

For the language and visual detector ($\vilangOOP$), we concatenate the pooled output \{cls\} token from BERT with the output query embedding corresponding to the detected object from deformable-DETR, and give this concatenated embedding to a three-layer fully-connected classifier to predict in or out-of-place. We train the classifiers using known labels of in or out-of-place from our mess up algorithm.   

For the BERT-only model, we give the pooled output \{cls\} token from BERT as input to our classifier. For the visual-only model, we give the output query embedding corresponding to the detected object from deformable-DETR to the classifier. 

We use the same hyperparameters for training all classifiers. We use a batch size of 25, an AdamW optimizer with a learning rate of 2e-7 and weight decay of 0.01, and train for 20k iterations.

\subsection{Object centroid relations} \label{sec:rel_cent}
As described in Section 3.2 of the main paper, we define a set of three relations based on the estimated centroids of the detected objects within the scene. We use these relations for building our input to the BERT out-of-place detector. These relations are computed with the following metrics: 

(i) \emph{Supported-by}: A receptacle is defined as a type of object that can contain or support other objects. Sinks, refrigerators, cabinets, and tabletops are some examples of receptacles. For the floor receptacle class, we consider the point directly below the object at the height of the floor (lowest height in our map). For all centroids $C_t^{\textrm{rec}}$ corresponding to receptacle classes $L_t^{\textrm{rec}} \subseteq L_t$, we define the single object $L^{\textrm{supp}} \in L_t^{\textrm{rec}}$ that supports the detected $C^{det}$ object as:
\begin{equation}
    L^{supp} = \arg \min(D(C^{\textrm{det}}, C_{t;\textrm{ydiff}<0}^{\textrm{rec}}))
\end{equation}
Where $D(x,Y)$ is the euclidean distance between centroid $x$ and each centroid in $Y$, and $\textrm{ydiff}<0$ takes all tracked centroids which are below the height of the detected centroid.

(ii) \emph{next-to}: We define the objects $L^{\textrm{next}}$ that are next to the detected $C^{\textrm{det}}$ object as:
\begin{equation}
    L^{next} = D(C^{det}, C_t) < d
\end{equation}
Where $D(x,Y)$ is the euclidean distance between centroid $x$ and all centroids $Y$, and d is a distance threshold.

(ii) \emph{closest-to}: We define the single object $L^{\textrm{closest}}$ that is closest to the detected $C^{det}$ object as:
\begin{equation}
    L^{\textrm{closest}} = \arg \min(D(C^{\textrm{det}}, C_t))
\end{equation}
Where $D(x,Y)$ is the euclidean distance between centroid $x$ and all centroids $Y$.

\subsection{Relational graph convolutional network} 
As described in Section 3.3 of the main paper, we use a relational graph convolutional network to predict plausible receptacle classes for the out-of-place object. The memex graph nodes are the sum of a learned object category embedding and visual features obtained from cropping the deformable-DETR backbone with the object's bounding box at the closest navigable location to the object. We connect nodes in the memory graph by computing their relations as described in Section~\ref{sec:mem_rel}. For the out-of-place object node, we similarly sum the learned embedding of the object's category label and visual features obtained from cropping the deformable-DETR backbone with the detected bounding box. The scene graph nodes are deformable-DETR output query features in the initial mapping of the scene for all detections above a confidence threshold. We include a map type node which is initialized with a learned embedding for each of the four room types. 

We use the rGCN to message pass 1) within the memory graph, and 2) to bridge the memory, scene, and out-of-place nodes. Let $\nOOP$ denote the node of the out-of-place object initialized with a learned category class embedding and visual features.  


Following the rGCN formulation in~\cite{schlichtkrull2018modeling}, we first update the nodes in the memory graph to distribute information within the memory:
\small
\begin{equation}
    h_i^{(l+1)} = \sigma(\sum_{r \in \mathcal{R}^{mem}}\sum_{j \in \mathcal{N}^{mem}_{i,r}} \frac{1}{c_{i,r}} W_r^{(l)}h_j^{(l)} + W_0^{(l)}h_i^{(l)}),
\end{equation}
\normalsize
where $h_i^{(l)} \in \mathbb{R}^{d^{(l)}}$ is the hidden state of node $v_i$ in the l-th layer of the neural network, with $d^{(l)}$ being the dimensionality of this layer’s representations, $\mathcal{N}^{mem}_{i,r}$ denotes the set of memory neighbor indices of node i under relation $r \in \mathcal{R}^{mem}$, and $c_{i,r}$ is a problem-specific normalization constant.

Inspired by~\cite{zareian2020bridging}, we then define a set of four bridging edges $\mathcal{R}^{bridge}$, one to connect $\nOOP$ to the updated memory nodes of the same object class, one to connect $\nOOP$ to all current scene nodes, one to connect $\nOOP$ to the room type node, and one to connect the the updated memory nodes to current scene nodes with the same category label. We then message pass via the bridging edges: 
\small
\begin{equation}
    h_i^{(l+1)} = \sigma(\sum_{r \in \mathcal{R}^{bridge}}\sum_{j \in \mathcal{N}^{bridge}_{i,r}} \frac{1}{c_{i,r}} W_r^{(l)}h_j^{(l)} + W_0^{(l)}h_i^{(l)}),
\end{equation}
\normalsize
where $\mathcal{N}^{bridge}_{i,r}$ denotes the set of bridge neighbor indices of the target node under bridge relation $r \in \mathcal{R}^{bridge}$.

We use four relational graph convolutional layers for each stage of message passing. Finally, we run the updated out-of-place object node through a classifier layer to predict a probability distribution over proposed receptacle classes to search for placing the target object. We optimize with a cross entropy loss using the object's ground truth receptacle label from the training scenes. 

\subsection{Memex graph} \label{sec:mem_rel}
We use 20 of the 80 training rooms to construct the memex graph. As described in section 3.3 of the main paper, the memex graph is a large graph of object nodes and relational edges that provide the relational graph convolutional network with exemplar context of object-object and object-scene relations. We obtain the ground truth category labels for the objects and use ground truth information from the simulator to obtain the relations \textit{above}, \textit{below}, \textit{next to}, \textit{supported by}, \textit{aligned with}, and \textit{facing}. The memex remains a constant graph throughout all remaining training and testing scenes. We use simulator ground truth information for convenience, but note that we could instead obtain the neural memex graph from human annotations of real-world houses. We compute \textit{above}, \textit{below}, \textit{next to}, and \textit{supported by} similar to Section~\ref{sec:rel_cent}, but instead use a distance metric on the 3D bounding boxes. For \textit{aligned with}, we check if the 3D bounding boxes have parallel faces. For \textit{facing}, we note that the back of an object usually carries more of its mass (e.g. the back of a sofa). Thus, we look at the mass distribution of the object within its 3D bounding box, and take the box face with the most of the point mass in its direction to be the back of the object. An object is facing a second object if the frustum of its front 3D bounding box face intersects the second object. We only consider facing for the following classes: \textit{Toilet, Laptop, Chair, Desk, Television, ArmChair, Sofa, Microwave, CoffeeMachine, Fridge, Toaster}.


\subsection{Visual search network} 
As described in Section 3.4 of the main paper, we use a visual search network to propose search locations conditioned on an object class. The input to the network is a 3D occupancy map $ \in \mathbb{R}^{C \times D \times H \times W}$ with $C=116$, $D=64$, $H=128$, $W=128$. $C=116$ represents a channel for each possible category in AI2THOR, as described in Section~\ref{sec:obj_track}. We first tile classes along all heights in $\mapthreeD$ to obtain a 2D input $ \in \mathbb{R}^{(C \cdot D) \times H \times W}$ to the network. This enters four 2D convolutional layers and returns a feature map $V^{uncond} \in \mathbb{R}^{C \times H \times W}$. The target object class is encoded with a learned category embedding and matrix multiplied with the feature map to condition the network on the target class. This is sent as input to four additional 2D convolutional layers to get a final output map $V^{cond} \in \mathbb{R}^{H \times W}$. We optimize this with a binary cross entropy loss on each 2D position independently using a Guassian-smoothed 2D map of ground truth object positions in the training scenes. Our output map provides spatial positions at a resolution of 128$\times$128. Since our output map need not predict a single location to search, we give positive samples significantly larger class weight than the negative samples to encourage high recall of the true location in the thresholded area. 

\section{Experimental details} \label{sec:exp_det}

\subsection{Tidying task} \label{sec:messup}
\begin{figure}[t!]
    \centering
    \includegraphics[width=\textwidth]{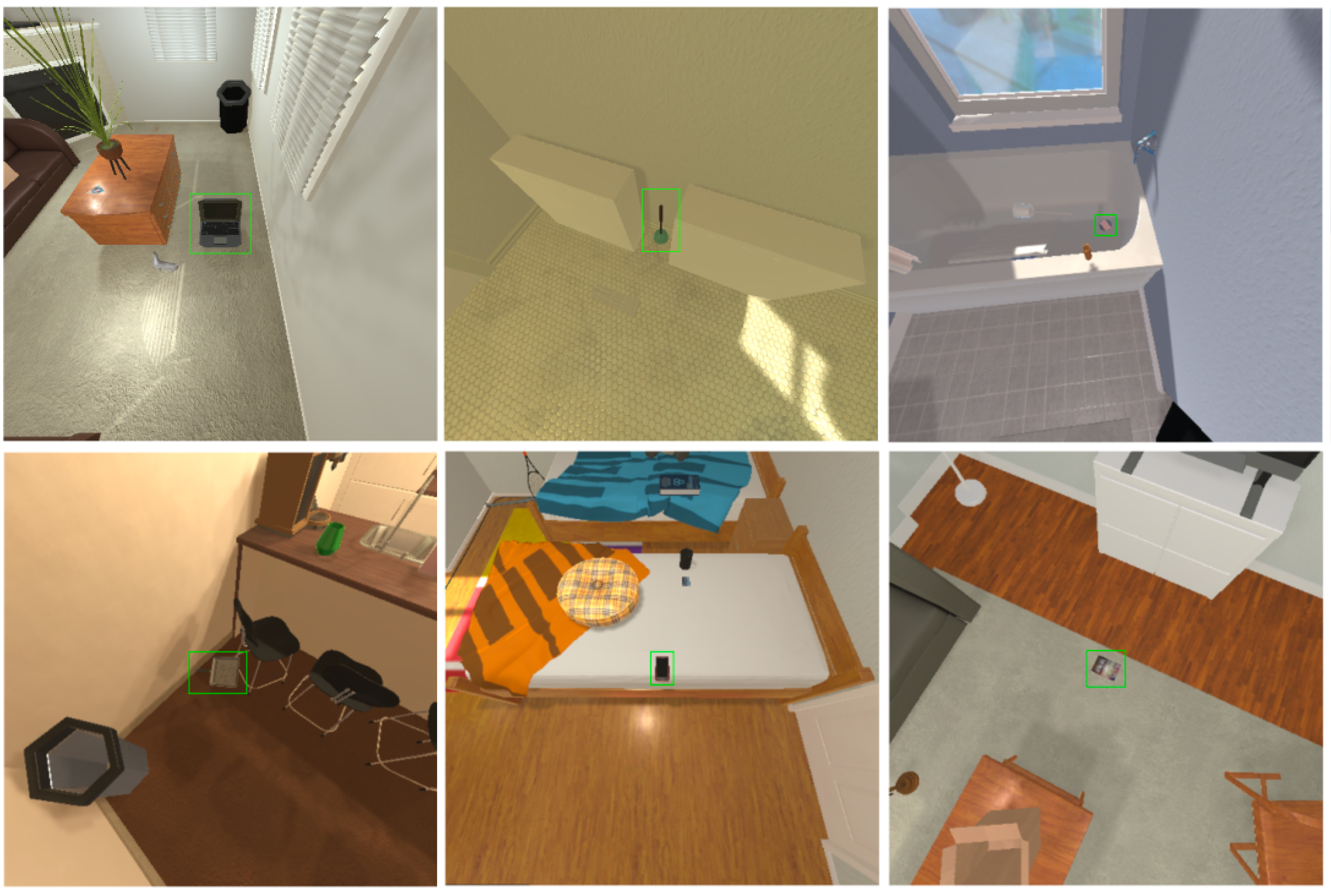}
    \caption{Example images of out-of-place objects.}
    \label{fig:oop_examples}
\end{figure}

Our tidying task begins with moving $N$ objects out of their natural locations in the scene. We use $N=5$ and generate five messy configurations per test room (total of 20 rooms $\times$ 5 configurations = 100 test configurations). For each object to be moved out-of-place, we randomly select a pickupable object, spawn an agent to a random navigable location in the scene at a random orientation in increments of 90 degrees, and with probability $p$, drop the object at the agent's location, or with probability $1-p$, throw the object with a constant force and let AI2THOR's physics engine resolve the final location (action "ThrowObject" in AI2THOR). We use $p = 0.5$. In AI2THOR, the throw distance of an object depends on its pre-defined mass, and thus the throw distance will change depending on the object. We keep the throw force constant at 150.0 newtons. We disable object breaking so that no objects are changed to their breaking state after dropping or throwing them. We show examples of out-of-place objects in Figure~\ref{fig:oop_examples}. 

We define an episode as the time from the spawn of the agent in the messy environment to the time the agent executes the ``done" action, or 1000 steps have been taken (whichever comes first). Once the tidying episode begins, the agent is spawned near the center of the map. At each time step, the agent is given an RGB and depth sensor, and its exact egomotion in terms of how far each action takes the agent and in what direction. During the out-of-place detection phase, $\model$ samples random locations within its 2D map to search. 

\subsection{Human placement evaluation} 
\begin{figure}[t!]
    \centering
    \includegraphics[width=\textwidth]{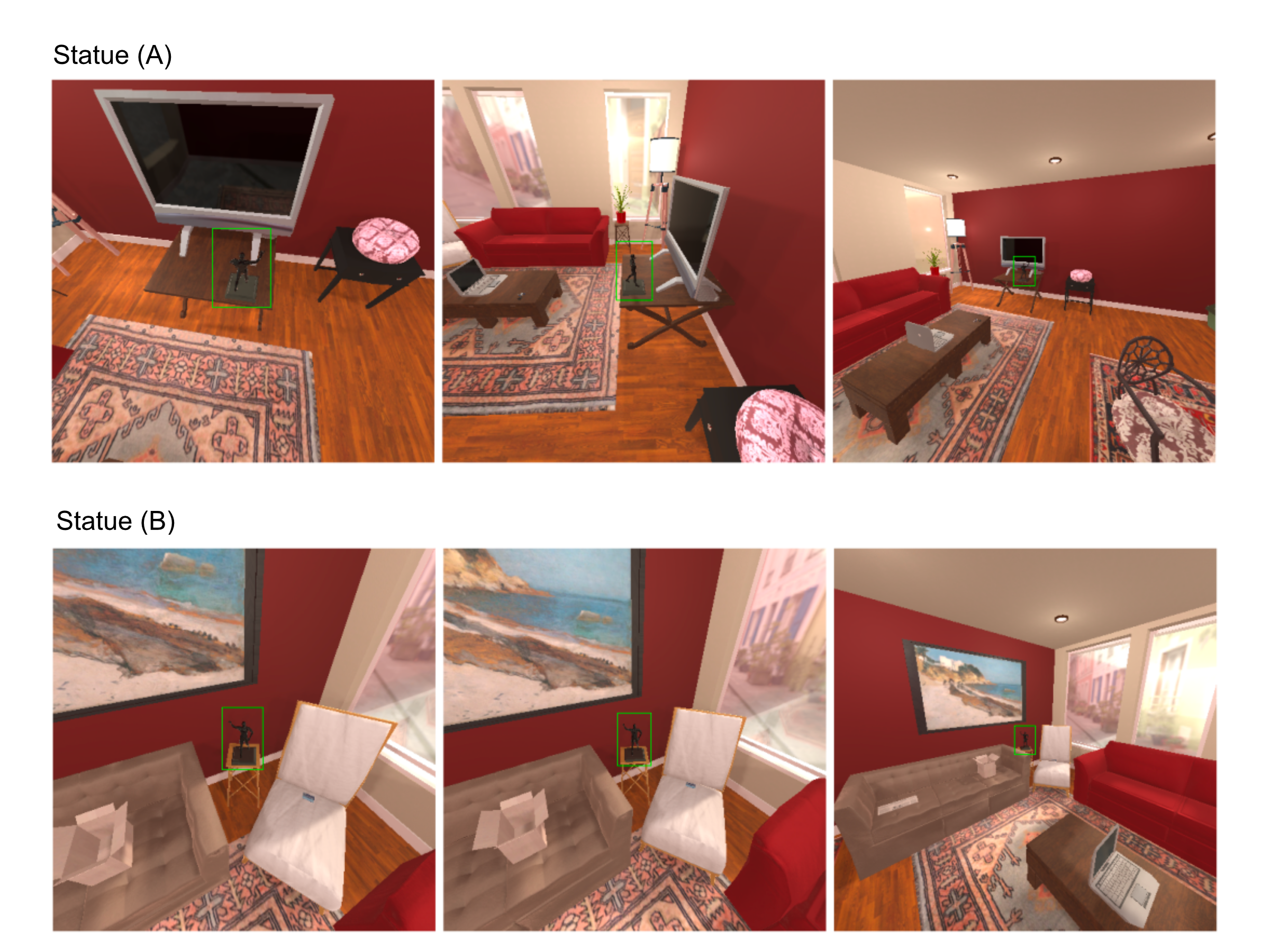}
    \caption{Example images shown to Amazon Mechanical Turk evaluators.}
    \label{fig:mechturk}
\end{figure}

We report in Section 4.2 of the main paper a human evaluation of $\model$ placements compared to baselines. We use the Amazon Mechanical Turk interface to query human evaluators as to whether they prefer $\model$ placements compared to baseline placements. For all successful placements by the agents, we generate three images of each placement to show the object from three distinct viewing angles, as shown in Figure~\ref{fig:mechturk}. We instruct the evaluators to choose between the placements of $\model$ and the baseline placement by looking at the images and picking which position of the object they would prefer. The full instructions given to the human evaluators for an example statue placement is displayed below. For this evaluation, we only consider objects which were picked up by both agents ($\model$ and the baseline). 

\small
\emph{Consider a scenario where you are putting the statue into its correct location in a room. Please choose which location you would prefer to place the statue within the room. The two options (A \& B) represent two different possible locations of the statue in the same room (in the images the location of the statue is shown with a box). Each option (A \& B) show the object from three distinct camera angles to help you make your decision. Important: Please judge only by the placement location of the object within the room, and NOT by the orientation of the object on the supporting surface.}

\subsection{Out-of-place detection evaluation} \label{sec:vsneval}
We evaluate the out-of-place detector performance in Section 4.3 on the same messy test scenes used for the tidying-up task. We generate 20 random views of each messy configuration where at least one out-of-place objects is in view. The total evaluation consists of 2000 images (20 scenes $\times$ 5 configurations $\times$ 20 views = 2000). We evaluate each detector by measuring average precision across all the images, where in and out-of-place are the two categories. 

\subsection{Exploration with visual search network evaluation} \label{sec:vsneval}
We evaluate the visual search network to assist in object goal navigation for objects in their default locations in the AI2THOR test scenes (20 scenes in total) in Section 4.4. For each test scene, the agent is tasked with finding each object category that exists at least once in the test scene. Each episode involves finding an instance of a given category. We consider all object categories across the AI2THOR simulator (116 categories). Tasking the agent under these specifications provides 591 total episodes in the evaluation. As mentioned in the main text, the agent is successful when the agent is within 1.5 meters of the target object and the object is visible to the agent. To declare success, the agent must execute the "Stop" command. If "Stop" is not executed within the maximum number of steps (200 max), the episode is automatically considered a failure and the next episode will begin. Both $\model$ and the baseline presented in Table 2 of the main text use the same object detector and navigation modules from Section 3.1 of the main paper. The only difference is how the model selects locations in the scene to search for the object-of-interest. For both $\model$ and the baseline, the agent executes the "Stop" command after the object category has been detected above a threshold and the agent has navigated to the detected object using the estimated 3D centroid. 

\subsection{Updating placement priors by instruction} \label{sec:nleval}
We show that we can alter the output of the language out-of-place detector by pairing specific language input with a desired label after additional training in Section 4.3. To do so, we first train the language detector ($\langOOP$) as described in Section~\ref{sec:oop_details} and Section 3.2 of the main paper. We then target a relation-label pairing. For example, we may want the relation "alarm clock supported-by the desk" to output the label "out-of-place" (which does not appear in the unaltered training set) whenever the relation occurs. Then, for an additional amount of (9k) iterations, whenever the relation "alarm clock supported-by the desk" appears in the training batch, we pair the sample with the "out-of-place" label as supervision. 

\section{Additional results} \label{sec:add_exp}

\subsection{2021 Rearrangement Challenge}
In section 4.5 of the main paper, we report the performance of $\model$ on the 2022 rearrangement benchmark. We additionally report performance on the 2021 rearrangement benchmark in Table~\ref{tab:rear_2021}. 

\begin{table}[h]
\centering
\caption{Test set performance on 2-Phase Rearrangement Challenge (2021).}
\label{tab:rear_2021}
\begin{tabular}{@{}lcccc@{}}
\toprule
& \%  FixedStrict $\uparrow$ & \% Success $\uparrow$ & \% Energy $ \downarrow$ & \% Misplaced $\downarrow$\\

 \midrule
 $\model$ & \textbf{8.9} & \textbf{2.6} & \textbf{93} & \textbf{95} \\
 $\model$ \scriptsize\textit{+noisy pose} & 6.6 & 1.9 & 97 & 98 \\
 $\model$ \scriptsize\textit{+est. depth} & 5.5 & 1.4 & 96 & 97 \\
 $\model$ \scriptsize\textit{+noisy depth} & 8.9 & 2.3 & 93 & 95 \\
 Weihs \textit{et al.}~\cite{RoomR} & 1.4 & 0.3 & 110 & 110 \\
\bottomrule
\end{tabular}
\end{table}

\subsection{Visualizations of the Visual Search Network}
\begin{figure}[t!] \label{}
    \centering
    \includegraphics[width=\textwidth]{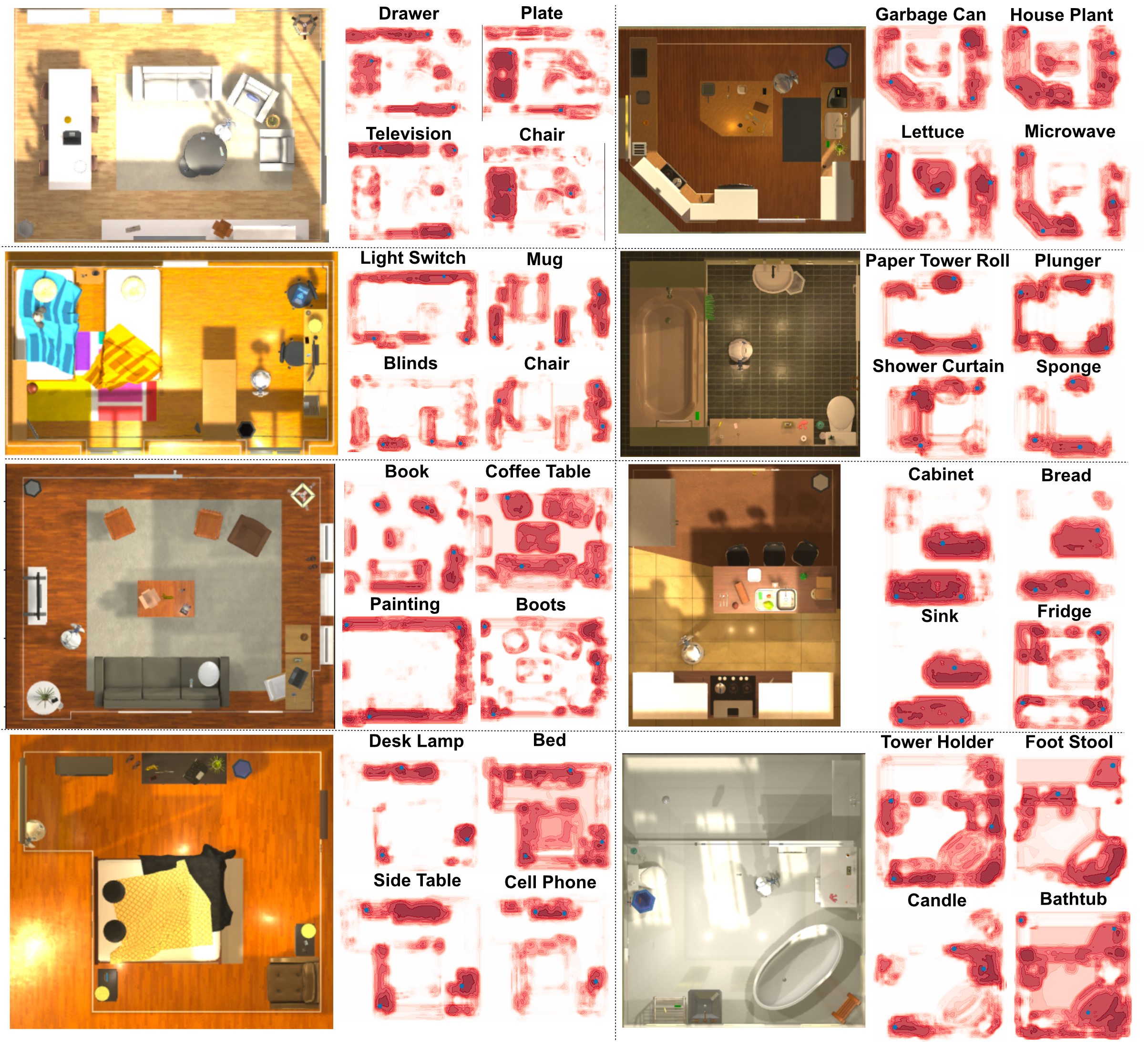}
    \caption{\textbf{Examples of the output of the Visual Search Network in test scenes.}}
    \label{fig:vsn_full_plot}
\end{figure}

In Section 4.4 of the main paper, we displayed visualizations of the Visual Search Network predictions. We provide additional visualizations of the sigmoid output of our Visual Search Network conditioned on an object category in test rooms in Figure~\ref{fig:vsn_full_plot}. We display an overhead view of the full scene on the left, and the network predictions corresponding to the overhead spatial locations on the right conditioned on four randomly-selected object categories. Darker red corresponds to higher probability. The blue dot indicators plotted in the prediction maps correspond to the search locations for the agent to visit after thresholding and farthest point sampling (for \# location = 3). The output generally puts the highest probability at plausible areas for the category to exist. However, occasionally the network puts high probability where it should not. For example, the network puts high probability near a dresser for category "Bed", or near the armchair for category "Coffee Table". This may be in part due to our training procedure to prioritize high recall over precision of the true location in our cross entropy weighting.

\subsection{Evaluation of altering priors with natural language}
In Section 4.6 of the main paper, we showed for a single example that we can alter the learned priors of the out-of-place detector using external language input. We augment training with nine additional object relation pairs that are among the most commonly found in the AI2THOR houses and pair the relation with an out-of-place label. The relation pairs include "alarm clock is supported by desk" (from main text), "Soap bottle is supported by countertop", "Pen is supported by desk", "Laptop is supported by desk", "Pillow is supported by bed", "Toilet paper is support by toilet", "salt shaker is supported by countertop", "Spatula is supported by countertop", "Statue is supported by shelf", and "Vase is supported by shelf". We follow the same training procedure as in Section~\ref{sec:nleval}. The average change in probability across test houses for examples where the relation appears is shown in Table~\ref{tab:nlmoretable}. The significant change in probability indicates we are able to change the detector output with simple language instructions. 
\begin{table}
\centering
\caption{\textbf{Altering priors with instructions.} 
The out-of-place confidence of the out-of-place classifier before and after augmenting training with the uncommon relation-label pairing.}
\label{tab:nlmoretable}
\begin{tabular}{@{}lccc@{}}
\toprule
& Before instruction & After instruction \\ \midrule
Alarm Clock supported-by Desk & .10 & .70 \\
Knife supported-by Dining Table & .44 & .91 \\
Bowl supported-by Dining Table & .23 & .71 \\
SoapBar supported-by Toilet & .21 & .68 \\
Laptop supported-by Bed & .25 & .71 \\
Apple supported-by CounterTop & .14 & .62 \\
Mug supported-by CounterTop & .27 & .77 \\
Newspaper supported-by Sofa & .43 & .98 \\
Pillow supported-by Bed & .56 & .70 \\
Book supported-by Desk & .63 & .88 \\
\bottomrule 
\end{tabular}
\end{table}

\end{document}